%% file: main.tex
\documentclass[10pt,twocolumn,letterpaper]{article}

\usepackage{iccv}              % To produce the CAMERA-READY version

\definecolor{iccvblue}{rgb}{0.21,0.49,0.74}
\usepackage[pagebackref,breaklinks,colorlinks,allcolors=iccvblue]{hyperref}

\def\paperID{12331} % *** Enter the Paper ID here
\def\confName{ICCV}
\def\confYear{2025}

\title{
$\bf{D^3}$QE: Learning Discrete Distribution Discrepancy-aware Quantization Error for Autoregressive-Generated Image Detection
}

\author{
Yanran Zhang$^{1,}$\thanks{Equal contributions. ~\textsuperscript{\dag}Corresponding authors.} ~~~
Bingyao Yu$^{1,*,\dagger}$  ~~~
Yu Zheng$^{1}$ ~~~
Wenzhao Zheng$^{1}$ ~~~
Yueqi Duan$^{2}$ ~~~
Lei Chen$^{1}$ ~~~\\
Jie Zhou$^{1}$ ~~~
Jiwen Lu$^{1,\dagger}$ \\
$^1$ Department of Automation, Tsinghua University, China\\
$^2$ Department of Electronic Engineering, Tsinghua University, China\\
{\tt\small \{zhangyr21\}@mails.tsinghua.edu.cn;} {\tt\small \{wenzhao.zheng\}@outlook.com;} \\{\tt\small \{yuby, yu-zheng, duanyueqi, leichenthu, jzhou, lujiwen\}@tsinghua.edu.cn} \\
}

\begin{document}
\maketitle
\input{sec/0_abstract}    
\input{sec/1_intro}

\input{sec/2_related_work}
\input{sec/3_Methods}

\input{sec/4_Experiments}

\input{sec/5_Conclusion}
\input{sec/6_acknowledgement}

{
    \small
    \bibliographystyle{ieeenat_fullname}
    \bibliography{main}
}

\end{document}

%% file: sec/0_abstract.tex
\begin{abstract}
The emergence of visual autoregressive (AR) models has revolutionized image generation while presenting new challenges for synthetic image detection.
Unlike previous GAN or diffusion-based methods, AR models generate images through discrete token prediction, exhibiting 
both marked improvements in image synthesis quality and unique characteristics in their vector-quantized representations. 
In this paper, we propose to leverage \textbf{D}iscrete \textbf{D}istribution \textbf{D}iscrepancy-aware \textbf{Q}uantization \textbf{E}rror (\textbf{D$^3$QE}) for autoregressive-generated image detection that exploits the distinctive patterns and the frequency distribution bias of the codebook existing in real and fake images. 
We introduce a discrete distribution discrepancy-aware transformer that integrates dynamic codebook frequency statistics into its attention mechanism, fusing semantic features and quantization error latent.
To evaluate our method, we construct a comprehensive dataset termed \textbf{ARForensics} covering 7 mainstream visual AR models.
Experiments demonstrate superior detection accuracy and strong generalization of \textbf{D$^3$QE} across different AR models, with robustness to real-world perturbations. Code is available at \href{https://github.com/Zhangyr2022/D3QE}{https://github.com/Zhangyr2022/D3QE}.
\end{abstract}

%% file: sec/1_intro.tex
\section{Introduction}
\label{sec:intro}
With the advent of Generative Adversarial Networks (GANs)~\cite{GoodfellowPMXWOCB14GAN} and Variational AutoEncoders (VAEs)~\cite{KingmaW13VAE}, significant advancements have been made in the realm of generative AI technology within computer vision. Following this, the emergence of innovative technologies such as Flow Models and Diffusion Models~\cite{song2020denoising,RombachBLEO22SD} has further enhanced the fidelity and quality of image generation. Nowadays, Autoregressive Models~\cite{tian2024VAR,Sun24LlamaGen,Han24Infinity} are capable of not only accurately capturing the structural features of images but also efficiently producing high-quality visual content.
On one hand, visual generative models have the potential to drastically reduce the time expenditure associated with manual creation, thereby empowering industries such as art, film production, and education. On the other hand, while these models facilitate the easy acquisition of images that can be indistinguishable from reality to the human eye, they also usher in a host of potential social risks and ethical dilemmas.
\begin{figure}[t]
    \centering
    \includegraphics[width=\linewidth]{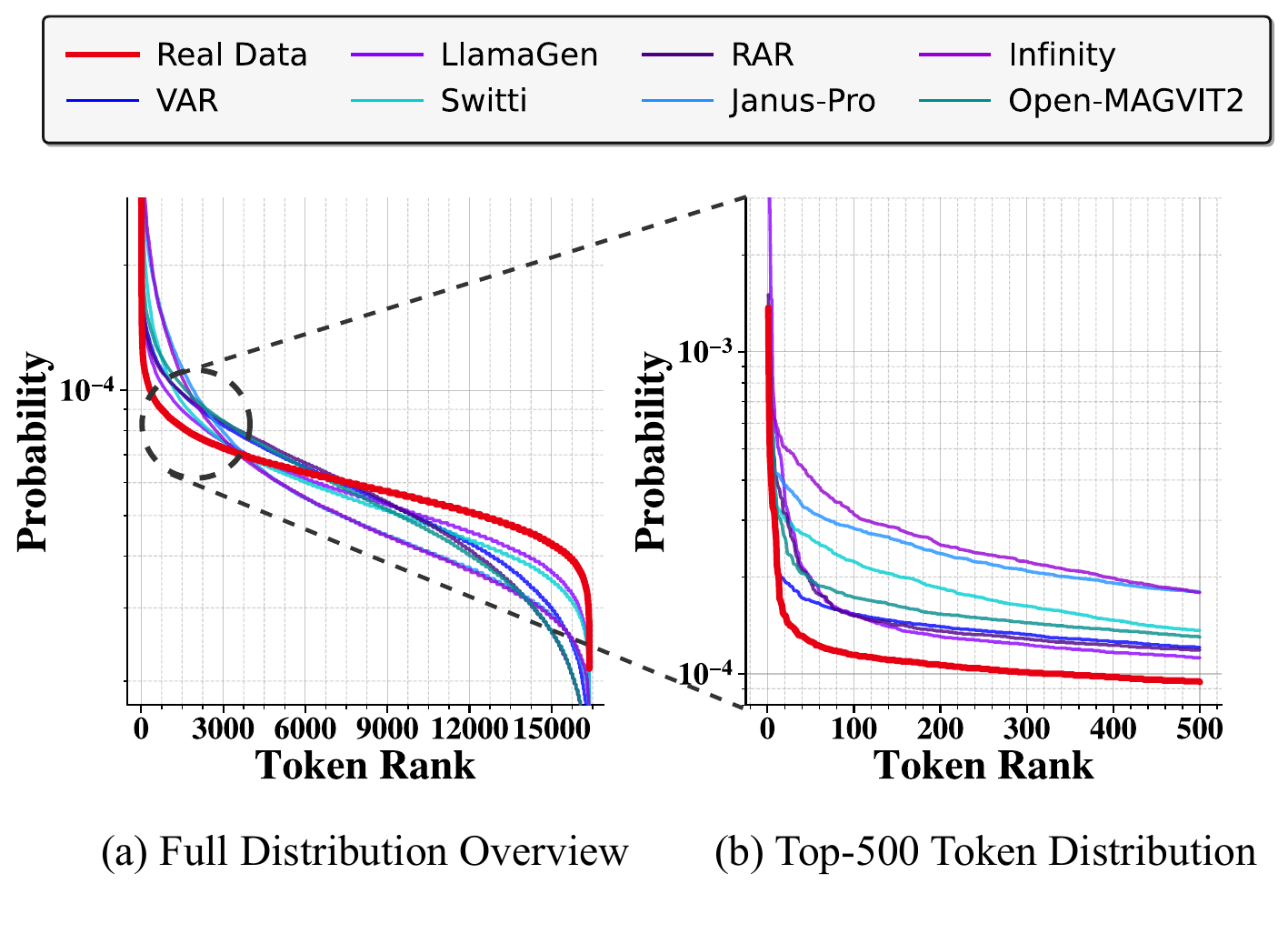} 
 \vspace{-30pt}
    \caption{\textbf{Visualization of Discrete Distribution Discrepancy.} To elucidate the mechanism of \textbf{D$^3$QE}, we analyze token probability distributions from LlamaGen's tokenizer using autoregressive sampling. (a) shows the full codebook vector probability distribution, while (b) displays the top-500 activation probabilities. The real data exhibits pronounced long-tail characteristics, whereas generated samples demonstrate concentrated probability mass in the peak regions, which \textbf{D$^3$QE} leverages for detection.}
    \vspace{-20pt}
    \label{fig:intro}
\end{figure}

In recent times, considerable efforts have been dedicated to the detection of generated images, with the goal of mitigating the trust crisis and addressing privacy risks that arise from the use of generative models.
Existing detection methods have primarily focused on high-frequency artifacts in GANs~\cite{frank2020leveraging,qian2020f3net} or iterative noise patterns in diffusion models~\cite{wang2023dire}, overlooking the unique characteristics of autoregressive models' discrete encoding. Traditional detection methods based on superficial statistical features struggle to identify these samples because the artifacts manifest in the discrete latent space rather than in pixel-level patterns, making them particularly challenging to detect through conventional image analysis techniques. 
For the newly emerging generative models, the generalization capability of detectors is paramount, and the latest AR models present a significant challenge to their effectiveness.

Discrete coding enhances the efficiency of reasoning and fosters diverse outcomes in generative models, while also highlighting the variations in statistical distributions across different images. As illustrated in Figure~\ref{fig:intro}, the discrete feature reveals distinct patterns in the utilization of codebook tokens among various generative models, with a more pronounced discrepancy between real and generated images.
Motivated by this, we delve into the prior knowledge of codebooks to construct robust features and enhance their expressiveness. Furthermore, by integrating the frequency disparity between real and fake codebooks into the cross attention mechanism and aligning it with quantization error, we merge the features with the semantic features extracted by the backbone network. A classifier is then employed to predict the final outcome.
For the first time, we have established a new benchmark termed \textbf{ARForensics} for the detection of images generated by AR models, encompassing the current top-performing mainstream AR models.  Effectiveness and generalization of our method were rigorously tested in a challenging experimental setting that included GANs, diffusion models, and AR models, demonstrating its robust performance.

%% file: sec/2_related_work.tex
\section{Related Work}

\noindent\textbf{Visual Generation.}
In recent years, visual generation models have experienced rapid development, with generated images and videos finding widespread applications in creative design and media production~\cite{RombachBLEO22SD,sora}. Mainstream generative models encompass four paradigms: GANs, VAEs, Diffusion Models, and Autoregressive Models. GANs~\cite{GoodfellowPMXWOCB14GAN,RadfordMC15DCGAN} generate realistic images through adversarial training, with subsequent improvements~\cite{KarrasLA19ProGAN,BrockDS19BigGAN,KarrasLA19StyleGAN} significantly enhancing generation quality. VAEs~\cite{KingmaW13VAE,HigginsMPBGBML17betaVAE} are based on latent space, while follow-up studies~\cite{VahdatK20NVAE,OordVK17VQVAE,LeeKKCH22RQVAE} addressed the blurry reconstruction issue through improved encoding structures. Diffusion models~\cite{HoJA20diffusion,song2020denoising,NicholD21ImprovedDiffusion,0011SKKEP21ScoreDiffusion,RombachBLEO22SD,peebles2023scalable} achieve high quality generation through iterative denoising processes.
Recently, visual autoregressive models~\cite{OordKK16PixelRNN,OordKEKVG16PixelCNN} have demonstrated remarkable capabilities by discretizing visual content into sequences and progressively predicting conditional probabilities, offering advantages in training stability and generation speed. Related works include token-based autoregressive modeling~\cite{EsserRO21VQGAN,yu2023magvit,Sun24LlamaGen,Qu24tokenflow,hong2022cogvideo} and scale-based autoregressive modeling~\cite{tian2024VAR,Han24Infinity}, both achieving significant progress in visual content generation. With the rapid development of autoregressive models, exploring effective detection methods for autoregressive-generated images has become particularly crucial.

\vspace{6pt}
\noindent\textbf{AI-generated Image Detection.}
With the rapid advancement of generative models, AI-generated image detection techniques have become crucial for ensuring information security and maintaining digital media authenticity. Recent research has evolved from local feature analysis to global semantic mining. Early studies~\cite{mccloskey2018detecting,mccloskey2019detecting,nataraj2019detecting} focused mainly on handcrafted features, including color distribution anomalies, saturation differences, and texture co-occurrence patterns. However, these methods showed limited generalization to newer generative models. 

The research community has proposed numerous detection methods targeting specific generative architectures. For GAN detection, studies have shifted towards frequency domain analysis. For instance, CNNSpot~\cite{wang2020cnn} enhanced cross-GAN architecture generalization through optimized data augmentation, while FreDect~\cite{frank2020leveraging} revealed artifacts introduced by GAN upsampling operations in the frequency domain. Following the rise of diffusion models, UnivFD~\cite{ojha2023towards} leveraged ViT's pre-trained features to train universal linear classifiers, while DIRE~\cite{wang2023dire} and AEROBLADE~\cite{AEROBLADE} achieved detection based on ADM~\cite{dhariwal2021diffusion} reconstruction errors and autoencoder reconstruction errors, respectively. NPR~\cite{NPR} designed a detection network that targets artifacts from common upsampling operations, and FatFormer~\cite{liu2024forgery} integrated local forgery traces through CLIP adapters.

However, existing approaches face two core challenges. First, while most research focuses on GANs and diffusion models, specific detection methods for autoregressive generative models remain underexplored. Artifacts from these models may exist in directional correlations or latent space discretization features. Second, current benchmark datasets lack samples from autoregressive models, limiting the validation of generalization capabilities of detection methods. Although the Chameleon benchmark~\cite{yan2024sanity} has improved in terms of diversity and realism, a more comprehensive evaluation framework is needed to support research on emerging models.

\begin{figure*}[t]
    \centering
    \resizebox{\textwidth}{!}{
    \includegraphics[width=0.9\textwidth]{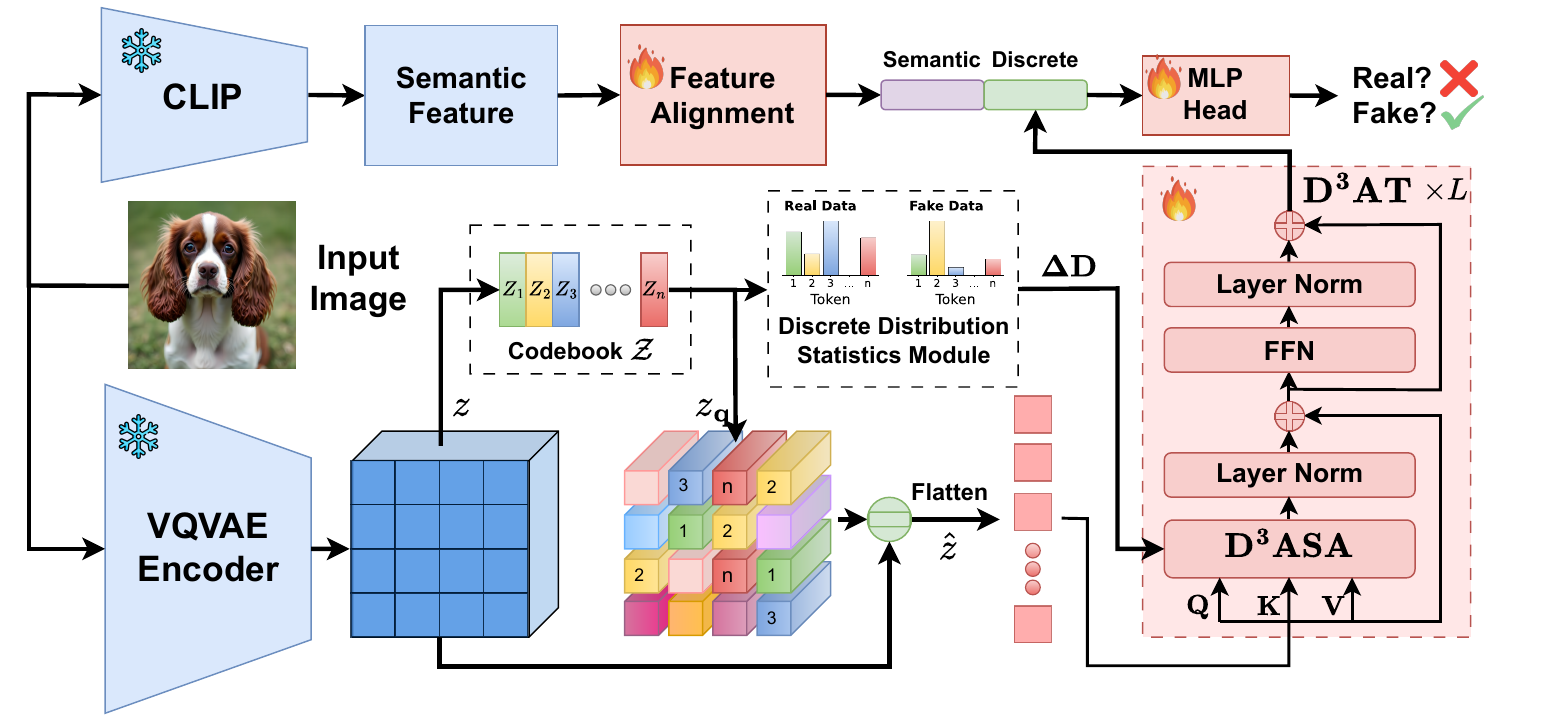} % 替换为你的流程图文件
    }
    \vspace{-20pt}
    \caption{\textbf{D$^3$QE pipeline.} Our approach first extracts quantized representations through a VQVAE encoder, computes the discrete distribution discrepancy between pre- and post-quantization features, and obtains discrete features via the \textbf{D}$^3$\textbf{AT} module. Semantic features are extracted using CLIP in parallel. The feature alignment module processes global semantic features, which then fuse with local discrete features for binary classification between generated and real samples. Blue snowflake symbols {\footnotesize\color{blue} \faSnowflake} indicate frozen parameters, while red flame symbols {\footnotesize\color{red} \faFire} denote trainable modules.
    }
     \vspace{-15pt}
    \label{fig:pipeline}
\end{figure*}

%% file: sec/3_Methods.tex
\section{Methods}

In this section, we present \textbf{D$^3$QE}, a novel framework for detecting autoregressive generated images. Our method leverages the unique discretization characteristics of visual autoregressive models. We first analyze the theoretical foundations of autoregressive modeling. Then, we detail our detection approach that combines discrete distribution awareness with semantic understanding.

\subsection{Preliminary}
\label{sec:ar_modeling}

\noindent\textbf{Visual Autoregressive Modeling.} 
Visual autoregressive models generate visual content in a sequential manner. These models operate through two key processes: discrete quantization and autoregressive modeling. The approach first trains a discrete variational autoencoder to quantize vectors in the latent space. Then, it performs autoregressive prediction of subsequent elements. This methodology has proven highly effective in capturing complex visual dependencies and generating high-quality content.~\cite{xiong2024autoregressive}

\noindent\textbf{Modeling via Next Token Prediction.} 
The next-token prediction methodology, borrowed from Natural Language Processing, has demonstrated remarkable generative capabilities in recent times.~\cite{EsserRO21VQGAN,Sun24LlamaGen}
At its core, this approach employs vector quantization through a VQVAE-like~\cite{OordVK17VQVAE} structure to compress continuous visual content into discrete sequences. The discretization process transforms input images into continuous latent representations, which are then quantized using a learnable codebook. After discretization, the model performs autoregressive prediction by estimating the probability of each subsequent token based on all preceding tokens. This sequential generation approach effectively captures both local patterns and global structural relationships in visual data.

\noindent\textbf{Modeling via Next Scale Prediction.}
VAR~\cite{tian2024VAR} pioneered the Next Scale Prediction approach, which discretizes content into multi-scale sequences through a hierarchical structure.
Unlike the token-by-token prediction, it models visual content from coarse to fine scales, where the autoregressive unit is a complete token map. The discretization process utilizes Residual Quantization from RQVAE~\cite{LeeKKCH22RQVAE}, obtaining discrete token maps through coarse-to-fine residual estimation. This hierarchical strategy enables high-quality image reconstruction with compact codebook capacity. The subsequent autoregressive modeling predicts finer-scale representations conditioned on preceding coarser scales.

\subsection{Motivation and Design Principles}

\noindent\textbf{Design Insights.} 
From the above analysis, we observe that discretization serves as a crucial component in mainstream visual autoregressive modeling. This discretization process fundamentally distinguishes visual autoregressive models from continuous generative paradigms (e.g. diffusion models), a design choice that has been systematically validated in seminal works such as VQVAE~\cite{OordVK17VQVAE}, VQGAN~\cite{EsserRO21VQGAN}, and VAR~\cite{tian2024VAR}. This architectural decision has profound implications for both model efficiency and generation quality.

Discretization has become a core feature of visual autoregressive models due to the following advantages. First, autoregressive models inherently decompose joint distributions through conditional probability chains. By transforming high-dimensional continuous visual data into discrete symbolic sequences, the model circumvents the curse of dimensionality while leveraging mature classification-based cross-entropy optimization paradigms from language modeling. Second, discrete distributions enable exact likelihood computation through classification cross-entropy loss, avoiding the mode collapse issues prevalent in continuous models. While continuous autoregressive models like MAR~\cite{LiTLDH24MAR} attempt to bypass discretization, they require diffusion losses for probability density estimation and often suffer from detail loss due to the smoothness of continuous latent spaces. Moreover, the structural homology between discrete tokens and NLP vocabularies enables visual autoregressive models to directly inherit architectural advantages from language models (as demonstrated in DALL·E~\cite{RameshPGGVRCS21DALLE} and Parti~\cite{YuXKLBWVKYAHHPLZBW22Parti}), facilitating unified cross-modal modeling.

\noindent\textbf{Overview.} 
Based on these insights, we propose to detect autoregressive generated images by analyzing their distinctive discrete distribution patterns. 
And our architecture consists of three key components: a quantization error representation module, a discrete distribution discrepancy-aware transformer, and a semantic feature embedding module.

\subsection{\textbf{D$^3$QE}}
\begin{figure}[t]
    \centering
    \vspace{-10pt}
    \includegraphics[width=\linewidth]{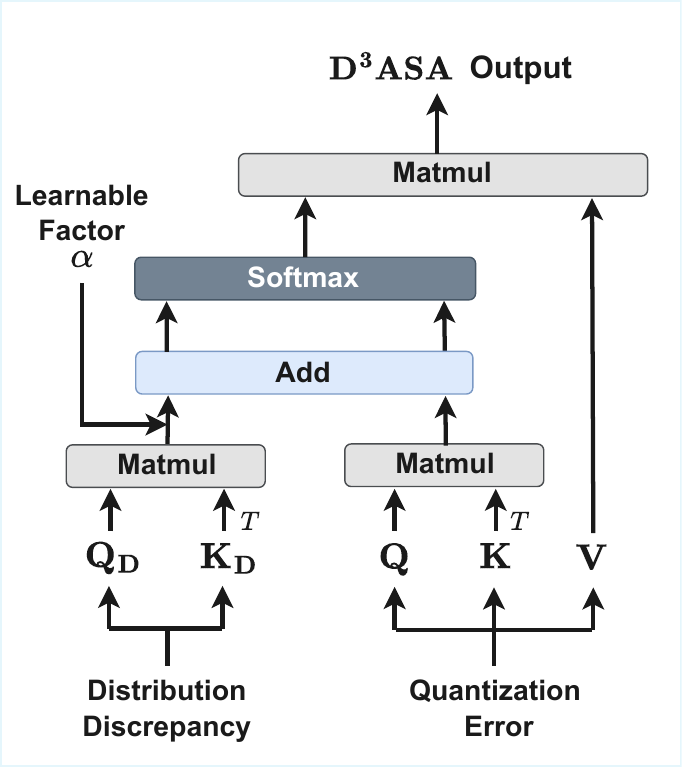} 
    \vspace{-25pt}
    \caption{Illustration of \textbf{D$^3$ASA} Module in Equation \ref{eq:D3ASA}, which incorporates distribution discrepancy information into the attention mechanism.}
    \vspace{-15pt}
    \label{fig:Attention}
\end{figure}

Based on the above analysis, we present our detection framework \textbf{D$^3$QE} that explicitly leverages the statistical signatures in autoregressive generated images.
The discretization process in visual autoregressive models introduces distinctive statistical signatures that can be leveraged for detection. This phenomenon occurs primarily because the finite codebook capacity struggles to fully capture the long-tailed distribution of natural images. The training objective of discrete VAEs forces the encoder to map similar features to the same codebook entries, resulting in high-frequency tokens corresponding to common local patterns. Rare patterns in real data such as specific object parts are compressed into high-frequency tokens due to their low occurrence rate, leading to reduced generation diversity. Furthermore, the explicit truncation introduced by top-p/top-k~\cite{SutskeverVL14Topk,SordoniGABJMNGD15Topk,HoltzmanBDFC20TopP} sampling strategies directly results in the truncation of long-tail distributions. As shown in Figure \ref{fig:intro}, these effects create observable differences in codebook distribution statistics between real and generated images.
To effectively capture these distinctive patterns between real and synthetic samples, we propose the following modules.

\noindent\textbf{Quantization Error Representation and Discrete Distribution Statistics Module.}
We first employ a frozen discrete autoencoder to tokenize images into discrete representations. Specifically, given an input RGB image $I \in {\mathbb{R}^{H\times W\times 3}}$, where $H$ and $W$ denote the height and width respectively, a deep neural network $\mathcal{E}$ encodes it into a continuous latent map $z=\mathcal{E}(I) \in {\mathbb{R}^{h\times w\times c}}$, where $h$, $w$, and $c$ represent the height, width, and channel dimensions of the latent space. The latent vectors are then projected into a learnable finite codebook $\mathcal{Z} = \{z_k\}_{k=1}^N \subset \mathbb{R}^c$, which contains $N$ discrete vectors. The process of finding the nearest codebook vector for each latent vector can be formulated as:
\begin{equation}
z_q = \left( \arg \min_{z_k \in \mathcal{Z}} \Vert {z}_{ij} - z_k \Vert \right) \in \mathbb{R}^{h \times w \times c}
\end{equation}
where ${z}_{ij}$ represents the latent vector at spatial position $(i,j)$ in the continuous latent map.

During training, we implement two discrete distribution tracking modules to monitor the distribution patterns of quantization indices for both real and synthetic images. These modules maintain frequency statistics for each codebook entry:
{\small\begin{equation}
D_s^{(t+1)}[k] = D_s^{(t)}[k] + \sum_{i,j} \mathbf{1}[q(z_{ij}) = k], s \in \{\text{real}, \text{fake}\}
\end{equation}}where $D_s^{(t)}[k]$ tracks the cumulative frequency of codebook index $k \in \{1,\dots,N\}$ at training step $t$, $q(z_{ij})$ denotes the index of the nearest codebook entry for the latent vector at position $(i,j)$.

After obtaining the quantized representation $z_q$, we compute the quantization error features to capture the discrepancy between continuous and discrete representations. This quantization gap potentially encodes distinctive patterns that differentiate real from synthetic images:
\begin{equation}
\hat{z} =( z_q - z)  \in \mathbb{R}^{h \times w \times c}
\end{equation}

%\vspace{6pt}
\noindent\textbf{Discrete Distribution Discrepancy-Aware Transformer (\textbf{D$^3$AT}).}
To effectively capture the distinctive patterns between real and synthetic samples, we propose a transformer-based module that explicitly incorporates codebook distribution information. Since the codebook usage patterns often differ significantly between real and synthetic images, we first compute their distribution discrepancy:
\begin{equation}
\mathbf{\Delta D} = \text{normalize}(D_{\text{fake}} - D_\text{real})
\end{equation}
where $\mathbf{\Delta D} \in \mathbb{R}^N$ represents the normalized difference in codebook entry frequencies.
The input features are reshaped into a sequence $\hat{\mathbf{X}}\in \mathbb{R}^{n\times c}$ following raster scan order, where $n$ denotes the sequence length and $c$ is the feature dimension. To effectively model both local and global dependencies while maintaining distribution awareness, our transformer architecture consists of $L$ layers. Each incorporating a novel Discrete Distribution Discrepancy-Aware Self-Attention (\textbf{D$^3$ASA}) mechanism. For the $\ell$-th layer:
\begin{align}
\hat{\mathbf{X}}_\ell &= \text{LN}(\text{\textbf{D$^3$ASA}}(\mathbf{X}_{\ell-1}, \Delta\mathbf{D})) + \mathbf{X}_{\ell-1} \\
\mathbf{X}_\ell &= \text{LN}(\text{MLP}(\hat{\mathbf{X}}_\ell)) + \hat{\mathbf{X}}_\ell
\end{align}
where $\text{LN}(\cdot)$ denotes layer normalization and $\mathbf{X}_0 = \hat{\mathbf{X}}$.
The \textbf{D$^3$ASA} mechanism enhances traditional self-attention by incorporating codebook distribution information. The distribution-aware attention is formulated as:
{\small
\begin{align}
\mathbf{Q_D} = \text{MLP}_q(\Delta\mathbf{D}), \quad  \mathbf{K_D}& = \text{MLP}_k(\Delta\mathbf{D}) \label{eq:qk} \\
\text{\textbf{D$^3$ASA}}(\mathbf{X}, \Delta\mathbf{D}) = \text{softmax}&\left(\frac{\mathbf{Q}\mathbf{K}^T}{\sqrt{d_k}} + \frac{\mathbf{Q_D}\mathbf{K}^T_\mathbf{D}}{\alpha}\right)\mathbf{V} \label{eq:D3ASA}
\end{align}
}where $\{\mathbf{Q}, \mathbf{K}, \mathbf{V}\}$ are query, key, and value matrices projected from input features $\mathbf{X}$, $\text{MLP}_q$ and $\text{MLP}_k$ are learnable distribution projections, and $\alpha$ is a learnable scaling factor that balances the contribution of distribution information.

\noindent\textbf{Semantic Feature Embedding.}
Beyond the distinctive patterns in local codebook token distributions, synthetic images often exhibit global semantic discrepancies compared to real images. To capture these high-level semantic differences, we leverage a pre-trained CLIP-ViT model to extract semantic features $\mathbf{F}_\text{CLIP}$. The CLIP features provide complementary global context information that helps identify subtle semantic inconsistencies in synthetic images.

\noindent\textbf{Classifier.}
Finally, we construct our classifier by combining both global semantic features and local token distribution patterns. To reduce computational complexity while preserving discriminative information, we first apply average pooling to the \textbf{D$^3$AT} output features to obtain a compact representation $\mathbf{F}_\text{D}$. The final prediction is computed by:
\begin{equation}
y = \text{MLP}(\text{concat}[\mathcal{A}_\text{D}(\mathbf{F}_\text{D}), \mathcal{A}_\text{CLIP}(\mathbf{F}_\text{CLIP})])
\end{equation}
where $\mathcal{A}_\text{D}$ and $\mathcal{A}_\text{CLIP}$ are feature alignment modules consisting of 
%lightweight 
MLPs and layer normalization to project features into a shared embedding space before concatenation.

%% file: sec/4_Experiments.tex
\section{Experiments}

In this section, we systematically constructed a comprehensive dataset of autoregressive model-generated images incorporating various generation strategies. Following dataset construction, we conducted systematic training and validation of our proposed model alongside existing SOTA baselines.
Through comparative analysis of model performance,
we validated the effectiveness of our proposed method and its critical components. We further evaluated the framework through  robustness and generalization experiments.

\begin{table*}[!ht]
  \caption{\textbf{ Performance comparison on ARForensics dataset.} Detection accuracy (Acc.) and average precision (A.P.) of various detectors (rows) against real and AI-generated images from different generative models (columns).}
 \vspace{-0.3 cm}
    \centering
    \renewcommand{\arraystretch}{1.2} 
    \setlength{\tabcolsep}{3pt} 
    \resizebox{\textwidth}{!}{ 
    \begin{tabular}{l c c c c c c c c c c c c c c | c c}
    \bottomrule \hline
       \multirow{2}*{\centering Method} & \multicolumn{2}{c}{LlamaGen}& \multicolumn{2}{c}{VAR}& \multicolumn{2}{c}{Infinity}& \multicolumn{2}{c}{Janus-Pro}& \multicolumn{2}{c}{RAR}& \multicolumn{2}{c}{Switti}& \multicolumn{2}{c}{Open-MAGVIT2}&  \multicolumn{2}{c}{Mean}\\
           & Acc. & A.P. & Acc. & A.P. & Acc. & A.P. & Acc. & A.P. & Acc. & A.P. & Acc. & A.P. & Acc. & A.P. & Acc. & A.P.\\ \bottomrule \hline
CNNSpot\cite{wang2020cnn}                 &99.94  & 99.94 &50.26  & 70.53 & 50.87 & 78.06 & 95.7 & 99.95  & 50.80 & 61.67 & 56.58 & 93.91 & 50.12 & 57.39 & 64.90 & 80.21 \\
FreDect~\cite{frank2020leveraging}                                            & 99.80 &  100.00 & 52.88 &  88.18 & 50.17 & 60.13 & 88.94 & 99.54 & 52.52 & 83.31 & 50.04 & 59.01 & 57.09 & 86.53  & 64.49 & 82.39 \\
Gram-Net~\cite{liu2020global}                                            & 99.57 & 99.98 & 55.04 & 84.57 & 52.38 & 76.80 & 74.48 & 97.33 & 49.95 & 52.72 & 57.74 & 88.66 & 50.08 & 53.72  & 62.75 & 79.11 \\
LNP~\cite{liu2022detecting}  & 99.48 & 99.99 & 49.64 & 55.42 & 49.76 & 49.94 & 99.53 & 99.98 & 49.69 & 55.61 & 70.28 & 94.16 & 49.63 & 54.92 & 66.86 & 72.86\\
UnivFD~\cite{ojha2023towards}  & 89.87 & 96.53 & 80.53 & 91.62 & 71.72 & 85.77 & 84.28 & 93.94 & 88.33 & 95.93 & 76.00 & 88.43 & 66.21 & 80.87 & 79.56 & 90.44\\
NPR~\cite{NPR}  & 99.96 & 100.00 & 56.87 & 88.68 & 88.48 & 97.98 & 93.67 & 99.18 & 52.30 & 74.99 & 51.97 & 87.04 & 63.00 & 92.11 & 72.32 & 91.43\\
  \rowcolor{gray!20} \textbf{D$^3$QE}(ours)   &  97.19 & 99.43 & 85.33 & 95.30 & 62.88 & 79.39 & 92.28 & 97.53 & 91.69 & 97.77 & 75.31 & 89.09 & 70.08 & 85.98  & \cred{82.11} & \cred{92.07} \\

\bottomrule
    \end{tabular}
}
  \label{tab:AR}
          \vspace{-0.3 cm}
\end{table*}
\subsection{Settings}

\noindent\textbf{ARForensics: A Dataset of Images Generated by Autoregressive Models.}
To validate the effectiveness of our method, we constructed the first benchmark dataset specifically designed for visual autoregressive models. We selected 7 representative autoregressive generative models --- LlamaGen~\cite{Sun24LlamaGen}, VAR~\cite{tian2024VAR}, Infinity~\cite{Han24Infinity}, Janus-Pro~\cite{chen2025janus}, RAR~\cite{yu2024randomized}, Switti~\cite{voronov2024switti}, and Open-MAGVIT2~\cite{luo2024open}, covering diverse architectures (token-based and scale-based) and resolutions. These models exhibit significant variations in their discretization processes and key technical parameters such as codebook capacity.

The dataset comprises 152,000 real samples and 152,000 generated samples. The real data come from ImageNet~\cite{deng2009imagenet}, one of the most influential benchmarks in computer vision, which contains manually annotated images across 1,000 fine-grained categories. ImageNet's rigorous quality control system provides a reliable foundation for model evaluation.
Our dataset consists of three splits: a training set of 100,000 LlamaGen-generated images paired with an equal number of randomly sampled ImageNet images (100 per category), a validation set of 10,000 image pairs, and a comprehensive test set incorporating 6,000 samples from each of the 7 autoregressive models, balanced with corresponding ImageNet test samples. It's worth noting that real images across all subsets are independently sampled to avoid evaluation bias from data overlap. This balanced design enables detection models to fully capture the characteristics of different generators while mitigating the impact of sample imbalance common in traditional datasets.

For image generation methodology, text-to-image models (Infinity, Janus-Pro, Switti) utilize a standard prompt template ``A photo of [class]", where [class] corresponds to ImageNet labels. Other autoregressive models (LlamaGen, VAR, RAR, Open-MAGVIT2) directly employ their ImageNet pre-trained versions, generating images through category-conditional synthesis. This approach produces synthetic images with high variability and reasonableness.

\noindent\textbf{Cross-Paradigm Test Set.}
To comprehensively validate our method's cross-domain generalization capability, we constructed a multi-modal test set of generated samples: 1) Based on the ForenSynths~\cite{wang2020cnn} dataset, we selected samples from representative GAN architectures including ProGAN~\cite{KarrasLA19ProGAN}, StyleGAN~\cite{KarrasLA19StyleGAN}, StyleGAN2~\cite{viazovetskyi2020stylegan2}, BigGAN~\cite{BrockDS19BigGAN}, CycleGAN~\cite{chu2017cyclegan}, StarGAN~\cite{choi2018stargan}, and GauGAN~\cite{park2019gaugan}; 2) We incorporated samples from mainstream diffusion models through the GenImage~\cite{GenImage} dataset, including ADM~\cite{dhariwal2021diffusion}, GLIDE~\cite{nichol2021glide}, Midjourney~\cite{midjourney}, Stable Diffusion V1.4~\cite{RombachBLEO22SD}, Stable Diffusion V1.5~\cite{RombachBLEO22SD}, and Wukong~\cite{wukong}, to evaluate adaptability of \textbf{D$^3$QE} across different generative paradigms.

\noindent\textbf{Evaluation Metrics.}
Our experiments strictly follow standard evaluation protocols in the field of generated image detection, employing Average Accuracy (Acc.) and Average Precision (A.P.) as core metrics. Acc is computed through binary classification with a fixed threshold of 0.5, while AP evaluates comprehensive performance of the classifier across different decision thresholds based on the area under the precision-recall curve.

\noindent\textbf{Baseline Methods.}
We conducted comparative experiments with state-of-the-art detection methods spanning multiple technical approaches: CNNSpot~\cite{wang2020cnn}, FreDect~\cite{frank2020leveraging}, Gram-Net~\cite{liu2020global}, LNP~\cite{liu2022detecting}, UnivFD~\cite{ojha2023towards}, and NPR~\cite{NPR}. All baselines were evaluated using official source code and recommended parameter configurations to ensure fair comparison.

\subsection{Implementation Details}
Our VQVAE encoder adopts the visual tokenizer in LlamaGen with a 16× downsampling tokenizer and a codebook of size 16,384. The encoder processes input images at $256 \times 256$ resolution. When optimizing subsequent modules, we freeze the CLIP encoder, VQVAE backbone, and codebook, while dynamically learning codebook statistics during training. The two-layer \textbf{D$^3$AT} module uses hidden dimension 512, while semantic features are extracted via CLIP-ViT from $224\times224$ preprocessed inputs. 
Experiments were conducted on an NVIDIA RTX 4090 GPU using PyTorch~\cite{paszke2019pytorch}. Training utilized AdamW via learning rate 0.0001, weight decay 0.01, batch size 32, for 10 epochs.

\subsection{Quantitative Results}
\begin{table*}[!ht]
  \caption{\textbf{Performance comparison on GAN-based synthesis using ForenSynths~\cite{wang2020cnn} test set.} Detection accuracy (Acc.) and average precision (AP) of various detectors (rows) against real and AI-generated images from different generative models (columns).}
 \vspace{-0.3 cm}
    \centering
    \renewcommand{\arraystretch}{1.2}
    \setlength{\tabcolsep}{4pt} 
    \resizebox{\textwidth}{!}{ 
    \begin{tabular}{l c c c c c c c c c c c c c c | c c}
    \bottomrule \hline
      %\multirow{3}*{Method} &\multicolumn{18}{c}{ Test Models}\\ 
       \multirow{2}*{\centering Method} & \multicolumn{2}{c}{ProGAN}& \multicolumn{2}{c}{StyleGAN}& \multicolumn{2}{c}{StyleGAN2}& \multicolumn{2}{c}{BigGAN}& \multicolumn{2}{c}{CycleGAN}& \multicolumn{2}{c}{StarGAN}& \multicolumn{2}{c}{GauGAN}&  \multicolumn{2}{c}{Mean}\\
           & Acc. & A.P. & Acc. & A.P. & Acc. & A.P. & Acc. & A.P. & Acc. & A.P. & Acc. & A.P. & Acc. & A.P. & Acc. & A.P.\\ \bottomrule \hline
CNNSpot~\cite{wang2020cnn}                 &50.26& 47.83&49.97& 43.89& 49.99& 46.49& 50.03& 41.16& 49.74& 50.56& 50.00& 44.66& 50.00& 52.73& 50.00& 46.76\\
FreDect~\cite{frank2020leveraging}                                            & 50.25&  66.83& 50.97&  71.46& 49.92& 56.13& 50.48& 55.12& 50.68& 53.87& 50.93& 98.44& 49.94& 33.03& 50.45& 62.12\\
Gram-Net~\cite{liu2020global}                                            & 49.78& 45.85& 50.04& 50.27& 49.77& 45.98& 49.78& 38.00& 48.07& 54.19& 50.00& 83.00& 50.00& 50.65& 49.64& 52.56\\
LNP~\cite{liu2022detecting}  & 50.00& 44.06& 50.69& 50.69& 50.01& 50.01& 50.00& 48.99& 50.00& 55.86& 50.00& 35.76& 50.00& 52.87& 50.10& 48.32\\
UnivFD~\cite{ojha2023towards}  & 88.17& 94.12& 72.98& 80.90& 72.23& 81.14& 88.78& 95.60& 71.23& 73.74& 79.99& 79.99& 91.52& 97.33& 80.70& 86.12\\
NPR~\cite{NPR}  & 51.36& 93.00& 52.54& 74.35& 50.93& 75.80& 50.30& 64.07& 48.83& 66.31& 53.83& 98.92& 50.03& 66.09& 51.12& 76.93\\
\rowcolor{gray!20} \textbf{D$^3$QE}(ours)    &  95.20& 97.68& 77.67& 88.65& 75.83& 88.61& 86.03& 94.79& 82.44& 92.31& 74.64& 85.65& 94.31& 97.94& \cred{83.73}& \cred{92.23}\\

\bottomrule
    \end{tabular}
}

  \label{tab:GAN}
          \vspace{-0.35 cm}
\end{table*}

\begin{table*}[!ht]
  \caption{\textbf{Performance comparison on diffusion-based generation using GenImage~\cite{GenImage} test set.} Detection accuracy (Acc.) and average precision (AP) of various detectors (rows) against real and AI-generated images from different generative models (columns).}
 \vspace{-0.3 cm}
    \centering
    \renewcommand{\arraystretch}{1.2}  
    \setlength{\tabcolsep}{6pt} 
    \resizebox{\textwidth}{!}{  
    \begin{tabular}{l c c c c c c c c c c c c | c c}
    \bottomrule \hline
       \multirow{2}*{\centering Method} & \multicolumn{2}{c}{ADM}& \multicolumn{2}{c}{Glide}& \multicolumn{2}{c}{Midjourney}& \multicolumn{2}{c}{SDv1.4}& \multicolumn{2}{c}{SDv1.5}& \multicolumn{2}{c}{Wukong}&  \multicolumn{2}{c}{Mean}\\
           & Acc. & A.P. & Acc. & A.P. & Acc. & A.P. & Acc. & A.P. & Acc. & A.P. & Acc. & A.P. & Acc. & A.P. \\ \bottomrule \hline
CNNSpot~\cite{wang2020cnn}                 &50.40& 55.54&54.81& 86.75& 50.93& 76.88& 50.23& 63.90& 50.29& 65.17& 50.35& 63.25& 51.17& 68.58\\
FreDect~\cite{frank2020leveraging}                                            & 51.83&  58.32& 63.82&  91.69& 50.57& 63.73& 56.80& 90.23& 56.73& 89.66& 55.75& 87.31& 55.91& 80.16\\
Gram-Net~\cite{liu2020global}                                            & 50.62& 50.54& 59.43& 90.96& 51.99& 78.01& 53.08& 82.31& 53.41& 82.46& 52.18& 77.37& 53.45& 76.94\\
LNP~\cite{liu2022detecting}  & 49.61& 55.52& 49.66& 54.10& 50.00& 51.08& 59.37& 88.02& 59.72& 88.45& 58.87& 87.51& 54.54& 70.78\\
UnivFD~\cite{ojha2023towards}  & 79.79& 90.86& 85.02& 94.07& 65.33& 78.21& 79.29& 91.16& 79.90& 91.01& 81.18& 92.16& 78.42& 89.58\\
NPR~\cite{NPR}  & 59.47& 69.62& 89.89& 98.39& 55.74& 97.38& 55.33& 89.98& 55.51& 90.38& 55.67& 75.19& 61.94& 86.82\\
\rowcolor{gray!20} \textbf{D$^3$QE}(ours)    &  70.43& 83.98& 88.89& 96.36& 61.21& 75.29& 83.33& 94.10& 83.37& 93.32& 84.43& 94.52&  \cred{78.61}& \cred{89.60}\\

\bottomrule
    \end{tabular}
}

  \label{tab:Diffusion}
          \vspace{-0.4 cm}
\end{table*}

In this section, we have thoroughly validated the effectiveness of our proposed method across a variety of generative models through extensive experiments. We conducted systematic experiments on 7 autoregressive models, 7 GAN models, and 6 diffusion models. The results demonstrate that our method achieves significant performance improvements compared to existing approaches across all models.

\subsubsection{Performance on ARForensics} As shown in Table \ref{tab:AR}, our proposed method demonstrates superior generalization capability across mainstream autoregressive models (including VAR, RAR, Open-MAGVIT2, etc.). Compared to traditional CNN-based detector CNNSpot, our method achieves significant improvements of 18.21\% and 11.86\% in average accuracy and average precision, respectively. In particular, for the latest scale-based autoregressive model VAR, our method achieves 85.33\% accuracy and 95.30\% AP, substantially outperforming the second-best method UnivFD at 80.53\%. This advantage stems from our deep modeling of discrete codebook statistical characteristics in autoregressive generation systems: capturing information loss during VQVAE compression through quantized residual features, while revealing codebook distribution concentration phenomena through codebook frequency difference maps.

Notably, previous baselines typically perform well on models that are architecturally similar to the training set model LlamaGen (e.g., Janus-Pro), but degrade significantly on architecturally distinct models (e.g., VAR/Switti), indicating their inability to capture common characteristics across autoregressive models. In contrast, our method maintains high detection performance on traditional raster-order models while demonstrating unique advantages in detecting both novel scale-based paradigms like VAR and random-scan-order models like RAR. These results validate that our \textbf{D$^3$AT} successfully captures cross-scale statistical biases through dynamic attention modulation. The features obtained from this module, when jointly optimized with CLIP semantic features, enable our codebook distribution-based framework to effectively handle  rapidly evolving autoregressive architectures.

\subsubsection{Performance on Cross-Paradigm Models}
To evaluate the generalization capability of our method, we directly apply our model trained on autoregressive samples to detect GAN and diffusion generated images. As shown in Tables \ref{tab:GAN} and \ref{tab:Diffusion}, our method demonstrates robust generalization across different generative architectures.

\noindent\textbf{Performance on GANs.} In GAN evaluation, our method achieves an average accuracy of 83.73\% and AP of 92.23\%, surpassing all baseline methods. Notably, we attain high AP of 97.68\% and 97.94\% on ProGAN and GauGAN, respectively. Despite GANs lacking explicit discretization, our codebook-based framework effectively captures distributional anomalies in GAN-generated images. This effectiveness likely stems from the hierarchical upsampling structure in GAN which imposes low-dimensional manifold constraints. The structure results in concentrated distribution patterns similar to discretization effects, which our \textbf{D$^3$AT} successfully identifies.

\noindent\textbf{Performance on Diffusion Models.} Our method demonstrates exceptional generalization to diffusion models, achieving an average accuracy of 78.61\% and AP of 89.60\%, comparable to state-of-the-art approaches. Detection accuracy reaches 83.33\%, 83.37\%, and 84.43\% on Stable Diffusion v1.4, v1.5, and Wukong respectively. While diffusion models' step-wise denoising fundamentally differs from autoregressive discrete generation, their iterative nature induces structured patterns in feature distributions. Our method's success in identifying these patterns can be attributed to the distribution-aware mechanism's sensitivity to similar feature patterns and the semantic embedding module's effective capture of global semantic inconsistencies.

\subsubsection{Robustness to Unseen Perturbations}
Images in real-world scenarios often undergo unpredictable perturbations, posing significant challenges for generated content detection. To validate the robustness of our method, we evaluated it on ARForensics datasets under JPEG compression (with quality $q\in[60,95]$) and center cropping (with crop factor $f\in[0.5,0.9]$ and subsequent resizing).
As shown in Figure \ref{tab:ablation}, experiments show that traditional methods generally suffer significant performance degradation under pixel-level perturbations, primarily due to the destruction of local artifact features left by generative models. In contrast, our approach demonstrates superior adaptability through discrete distribution awareness and feature fusion. Under JPEG compression, our method maintains detection AP above 85\% even when the quality factor drops to 60, consistently showing greater robustness than previous approaches. When facing severe cropping with $f=0.5$, our method still preserves over 80\% detection AP. These results validate the stability of our proposed multi-granularity feature fusion strategy across various perturbation conditions.

\begin{figure}[t]
    \centering
    \includegraphics[width=0.95\linewidth]{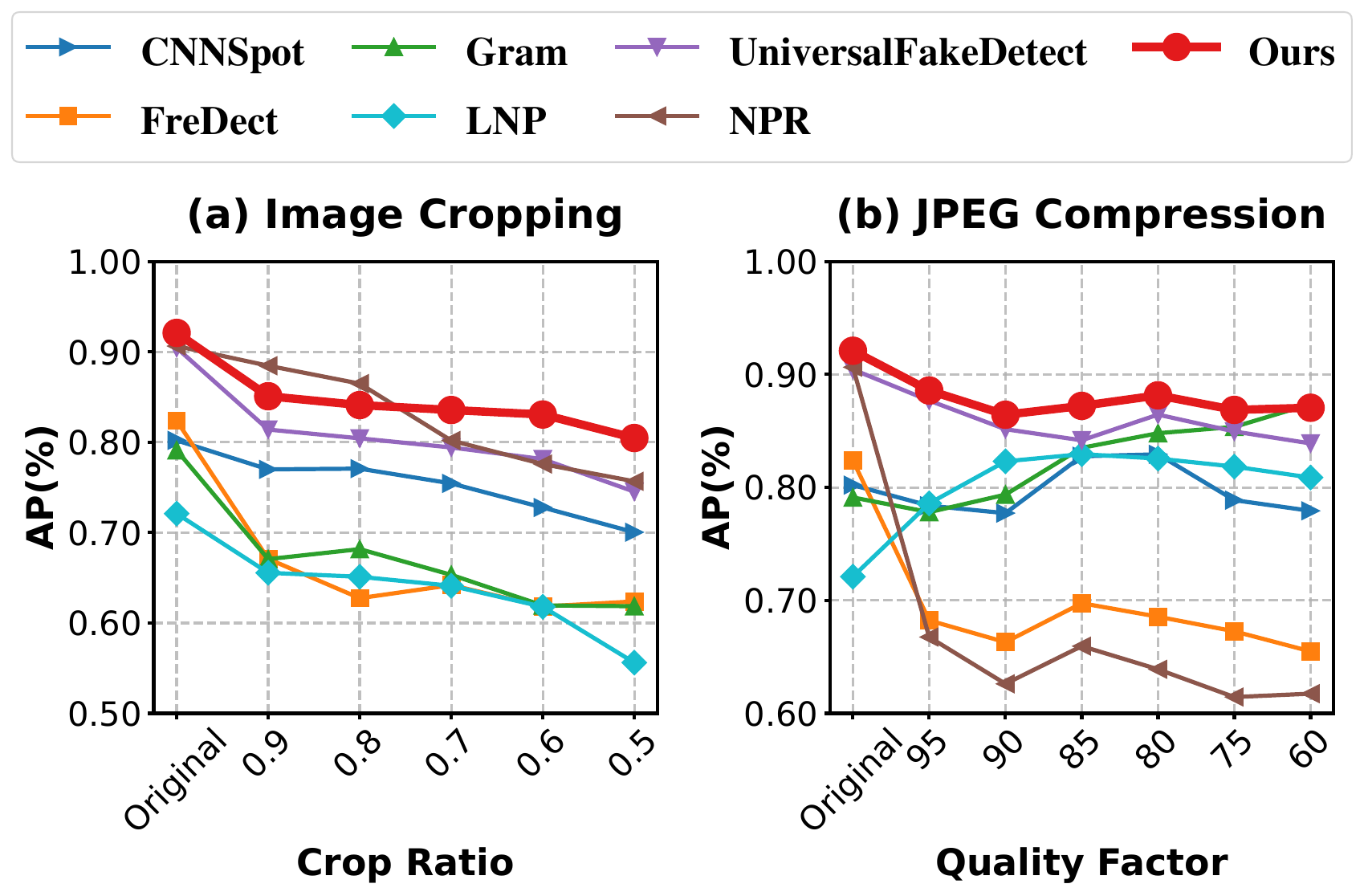} 
     \vspace{-10pt}
    \caption{\textbf{Robustness Analysis.} Performance comparison under image cropping and JPEG compression. Our method maintains superior accuracy across different perturbation levels, demonstrating strong robustness against common image transformations.}
    \label{fig:robustness_analysis}
     \vspace{-10pt}
\end{figure}
\subsection{Ablation Studies}

To validate the effectiveness of model components, we systematically analyzed how different module configurations and parameter settings affect detection performance.

In the Table \ref{tab:ablation}(a) of module analysis, the base model (Model {\ding{172} }) achieves 79.56\% accuracy using only CLIP semantic features. Incorporating VQVAE residual features (Model {\ding{173}}) improves performance to 79.92\%, demonstrating the effectiveness of discrete latent space modeling in enhancing feature representation.
Comparing discrete features $z_q$ (Model {\ding{174}}) with residual features $\hat{z}$ (Model {\ding{175}}) reveals that residual quantization information more precisely captures distributional shifts in generated images, improving accuracy by 0.33\%. Replacing the standard Transformer with our \textbf{D$^3$AT} (Model {\ding{176}}) further enhances detection accuracy to 82.11\%, validating the effectiveness of our codebook statistical feature fusion mechanism.

In the Table \ref{tab:ablation}(b) of dimension sensitivity tests, the \textbf{D$^3$AT} module achieves peak performance at 512 dimensions (82.11\%). Lower dimensions restrict representational capacity (80.83\% at 128 dimensions), while higher dimensions lead to overfitting (80.37\% at 1024 dimensions), indicating significant model sensitivity to feature dimensionality. These results demonstrate that the synergistic design of residual quantization features and discrete distribution discrepancy-Aware self-attention attention mechanisms is crucial to improve detection performance.
\begin{table}[!t]
\caption{\textbf{Ablation studies on model components and parameter settings. (R: Residual, D: Discrete, V: Vanilla)}}
\vspace{-0.3cm}
\label{tab:ablation}
\begin{subtable}[t]{0.615\columnwidth}
    \centering
    \caption{Module Analysis}
    \vspace{0.05cm}
    \renewcommand{\arraystretch}{1.2}
    \resizebox{\linewidth}{!}{
        \begin{tabular}{c|ccc|c}
        \bottomrule \hline
        \multirow{2}*{Model} & \multicolumn{3}{c|}{Module Configuration} & \multirow{2}*{Acc.} \\
        \cline{2-4}
        & CLIP & Latent & Transformer & \\ 
        \bottomrule \hline
        \ding{172}   & \checkmark & \xmark & \xmark & 79.56 \\
        \ding{173}   & \checkmark & R & \xmark & 79.92 \\
        \ding{174}  & \checkmark & D & V & 80.39 \\
        \ding{175}   & \checkmark & R & V & 80.72 \\
        \rowcolor{gray!20} \ding{176}   & \checkmark & R & \textbf{D$^3$AT} & 82.11 \\
        \bottomrule
        \end{tabular}
    }
    
    \label{tab:module_ablation}
\end{subtable}
\hfill
\begin{subtable}[t]{0.32\columnwidth}
    \centering
    \caption{Dimension Analysis}
    \vspace{0.05cm}
    \renewcommand{\arraystretch}{1.2}
    \resizebox{\linewidth}{!}{
        \begin{tabular}{c|c}
        \bottomrule \hline
        \textbf{D$^3$AT} dim & Acc. \\
        \bottomrule \hline
        128 & 80.83 \\
        256 & 81.57 \\
        384 & 80.95 \\
        \rowcolor{gray!20} 512 & 82.11 \\
        1024 & 80.37 \\
        \bottomrule
        \end{tabular}
    }
    \label{tab:dim_ablation}
\end{subtable}
\vspace{-0.5cm}
\end{table}
%%%%%%%%%%%%%%%%%%%%%%%
\begin{figure}[t]
    \centering
    \includegraphics[width=0.98\linewidth]{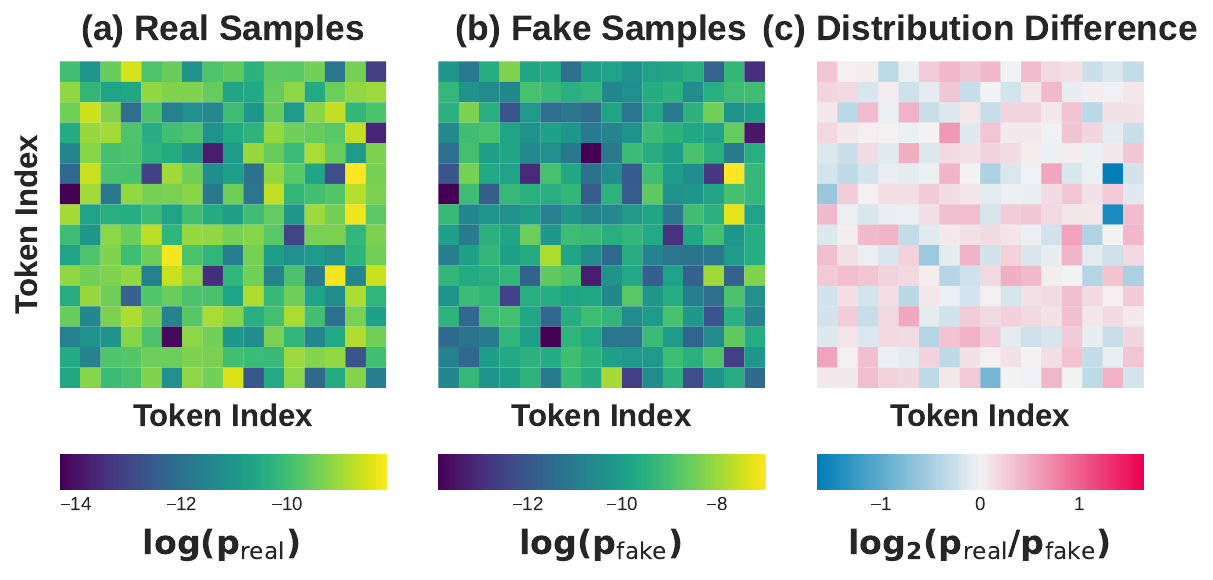} 
     \vspace{-11pt}
    \caption{\textbf{Visualization of codebook activation patterns.} Heatmaps show normalized logarithmic activation frequencies of VQVAE codebook vectors for (a) real samples and (b) generated samples, with (c) their log-ratio difference. Real samples exhibit uniform activation patterns, while generated samples show significant polarization in high-frequency regions.}
     \vspace{-16pt}
    \label{fig:token_analysis}
\end{figure}

\subsection{Qualitative Results}
To gain deeper insights into our model's intrinsic properties, we visualize the codebook activation distributions to reveal fundamental differences between real and generated samples. As illustrated in Figures \ref{fig:token_analysis}(a) and (b), we analyze the normalized logarithmic activation frequencies of the first 256 vectors in the VQVAE original codebook, based on the statistical distributions from our trained model.

The distributions reveal striking contrasts. Real samples exhibit balanced codebook utilization with uniform activation patterns. Generated samples, however, exhibit severe polarization. Their high-frequency codebook entries show anomalous peaks, with activation rates 3-5 times higher than real samples, while low-frequency regions show reduced coverage. The per-vector difference heatmap (Figure \ref{fig:token_analysis}(c)) illustrates this disparity, revealing mode collapse characteristics in the discrete latent space. These distributional patterns reflect inherent limitations of autoregressive models in capturing complex real-world distributions, providing strong empirical support for our detection framework.

%% file: sec/5_Conclusion.tex
\section{Conclusion}

In this paper, we have proposed to learn Discrete Distribution Discrepancy-aware Quantization Error (\textbf{D$^3$QE}) for autoregressive-generated image detection, which aims to exploit the distinctive patterns and the frequency distribution bias for various AR models. 
Further, we introduce a discrete distribution discrepancy-aware transformer to utilize dynamic codebook frequency statistics for combining semantic features with quantization error latent.
Finally, we construct a comprehensive dataset covering mainstream visual AR models to evaluate our method, and
experiments show that \textbf{D$^3$QE} achieves superior accuracy and strong generalization  across different AR models, while maintaining robustness under various real-world perturbations.

%% file: sec/6_acknowledgement.tex
\section*{Acknowledgement}
This work was supported in part by the National Nature Science Foundation of China under Grant 62441616, Grant 62336004, Grant 62406172, and Grant 62125603, in part by the China Postdoctoral Science Foundation No. 2023M741964, and in part by the Postdoctoral Fellowship Program of CPSF No. GZC20240841.